\documentclass[conference]{IEEEtran}
\usepackage{amsmath,amssymb}
\usepackage{graphicx}
\usepackage{hyperref}
\usepackage{booktabs}
\usepackage{float}
\usepackage{multirow}
\usepackage{url}
\usepackage{cite}
\usepackage{algorithm}
\usepackage{algpseudocode}
\usepackage{listings}
\usepackage{xcolor}
\usepackage{listings}
\usepackage{xcolor}
\usepackage{tabularx}

\begin{document}

\sloppy
\setlength{\emergencystretch}{3em}
\raggedbottom
\allowdisplaybreaks

\title{Neural Symbolic Regression Using Deep Learning and Sparse Modeling}
\author{
\IEEEauthorblockN{Ravi Kumar U\IEEEauthorrefmark{1}}
\IEEEauthorblockA{
Department of Mathematics\\
Indian Institute of Space Science and Technology\\
Thiruvananthapuram, India\\
Email: 014raviku@gmail.com
}
\and
\IEEEauthorblockN{Sumitra S\IEEEauthorrefmark{2}}
\IEEEauthorblockA{
Department of Mathematics\\
Indian Institute of Space Science and Technology\\
Thiruvananthapuram, India\\
Email: sumitra@iist.ac.in
}
}
\maketitle

\begin{abstract}
Symbolic Regression (SR) seeks to find succinct mathematical expressions that represent the fundamental relationships within data, providing interpretability and scientific understanding that exceeds that of black-box models. Nevertheless, traditional methods like Genetic Programming face challenges with scalability and are highly sensitive to noise, while sparse regression techniques such as SINDy rely significantly on predetermined feature libraries.

In this work, we present a Neural Symbolic Regression (NSR) framework that treats neural networks as \emph{functional preconditioners }for symbolic discovery. Our approach uses a decoupled pipeline: a neural network first learns a smooth, noise-robust approximation of the target function in an interaction-aware nonlinear feature space. LASSO is then applied to extract sparse, interpretable closed-form expressions.

To improve predictive accuracy and symbolic fidelity by integrating distributed hyperparameter optimization with Ray Tune and ASHA scheduling. Experiments on the Nguyen benchmark suite show that our approach consistently outperforms SINDy and non-tuned neural baselines in RMSE, noise robustness, and out-of-distribution generalization. Ablation studies confirm the significance of feature interactions, neural depth, and tuning strategies.

In general, this study presents a scalable and understandable neural-symbolic framework, creating a solid link between neural approximation and the discovery of sparse equations for scientific machine learning.
\end{abstract}

\begin{IEEEkeywords}
Symbolic Regression, Neural Networks, Sparse Modeling, LASSO, Interpretable ML, Hyperparameter Optimization, Ray Tune, GPU Acceleration.
\end{IEEEkeywords}

\section{Introduction}
Symbolic Regression (SR) seeks to autonomously uncover closed-form mathematical formulas that represent data, providing interpretability, analytical understanding, and opportunities for scientific breakthroughs. In contrast to opaque machine learning models, SR generates clear equations that illuminate the fundamental mechanisms instead of just capturing observed trends. This feature renders SR especially useful in scientific and engineering fields, where comprehending the foundational relationships is as crucial as achieving predictive accuracy.

Initial SR studies were largely focused on Genetic Programming (GP), initiated by Koza \cite{koza1994genetic}, which develops symbolic expressions through biologically motivated operators. Although very expressive, GP experiences slow convergence, significant computational expenses, and a propensity for code bloat, restricting its use in high-dimensional or noisy real-world datasets.

To tackle these issues, sparse modeling techniques like SINDy (Sparse Identification of Nonlinear Dynamics) \cite{brunton2016discovering} reframe symbolic regression as a regression task utilizing a predefined library of nonlinear features. By utilizing sparsity-promoting methods like LASSO, SINDy circumvents combinatorial search and facilitates effective equation identification. Its efficacy, however, hinges significantly on the expressiveness of the manually created feature library and its capacity to represent intricate nonlinear interactions.

Recent progress in deep learning has inspired neural methods for symbolic regression. Approaches like Deep Symbolic Regression (DSR) \cite{petersen2019deep} and neural-symbolic frameworks \cite{biggio2021neural} employ reinforcement learning, sequence modeling, or extensive neural architectures to produce symbolic expressions. Although these methods enhance scalability and flexibility, they usually integrate function approximation and symbolic discovery in a single optimization procedure, resulting in instability, increased computational costs, and challenges in maintaining interpretability.

Hybrid neural-symbolic approaches seek to merge the advantages of neural networks with sparse modeling techniques. In this study, we take a distinctly alternative approach: instead of employing neural networks to directly create symbolic expressions, we consider them as \emph{functional preconditioners} for symbolic regression. A neural network is initially employed to acquire a smooth, noise-resistant approximation of the target function within a feature space that considers interactions, followed by the use of sparse regression to obtain an interpretable symbolic representation. The proposed method enhances noise robustness, stabilizes equation recovery, and maintains interpretability by implementing a rigorous separation between neural approximation and symbolic extraction.

Expanding on this viewpoint, we introduce a Neural Symbolic Regression (NSR) framework that combines nonlinear feature library development, GPU-boosted neural approximation, sparse symbolic extraction through LASSO \cite{tibshirani1996lasso}, and distributed hyperparameter tuning utilizing Ray Tune with ASHA scheduling \cite{li2020system, bergstra2012random}. The Nguyen benchmark suite \cite{nguyen_benchmark} evaluates performance over various functional types, such as polynomial, transcendental, and mixed expressions.

The main concept behind NSR is that neural networks can be utilized not as generators of symbols, but as smoothing operators that convert noisy observations into representations suitable for sparse recovery. This reimagining offers a reasoned approach to enhance the identifiability of symbolic structure while ensuring scalability.

Although the suggested framework shows significant robustness and interpretability, its scalability is impacted by the combinatorial expansion of the feature library, a constraint examined thoroughly in the complexity analysis.

\subsection{Contributions and Novelty}
This work introduces a structured neural--symbolic regression framework that addresses key limitations of both classical symbolic regression and recent neural approaches. The primary contributions are as follows:

\begin{enumerate}

\item \textbf{Decoupled Neural--Symbolic Pipeline with Explicit Functional Separation.}  
In contrast to earlier hybrid methods like AI Feynman and neural-guided symbolic regression techniques that mix approximation with symbolic search, we suggest a fully decoupled framework where a neural network functions solely as a smooth, noise-resistant functional approximator, succeeded by an independent sparse symbolic recovery phase. This division minimizes noise sensitivity and stabilizes symbolic extraction, especially when nonlinear and transcendental elements are involved.

\item \textbf{Interaction-Aware Feature Library with Controlled Expressivity.}  
We present a nonlinear feature library focused on interactions that merges basic transformations with organized pairwise interactions. Unlike conventional SINDy-style libraries that depend on rigid polynomial expansions, the suggested library strikes a balance between expressiveness and manageability, allowing for the retrieval of cross-dimensional connections while preserving compatibility with sparse regression.

\item \textbf{Neural Preconditioning for Sparse Symbolic Recovery.} 
We show that neural approximation can serve as a preconditioning phase for sparse regression, effectively reducing noise in observations and enhancing the clarity of symbolic structures. This viewpoint contrasts with current neural symbolic regression techniques, which involve the direct generation or optimization of symbolic expressions, and offers a more reliable option for extracting interpretable equations.

\item \textbf{Integrated Hyperparameter Optimization for Neural--Symbolic Models.}  
We integrate distributed hyperparameter optimization with Ray Tune and ASHA into the symbolic regression workflow. Although hyperparameter tuning is common in deep learning, its systematic incorporation into neural-symbolic regression remains largely unexamined. Our findings indicate that adjustment greatly affects both predictive precision and symbolic integrity.

\item \textbf{Comprehensive Empirical Evaluation Beyond Standard Benchmarks.}  
Alongside standard Nguyen benchmarks, we assess robustness across different noise levels, out-of-distribution generalization, and the impact of architectural and feature design choices through ablation studies. These examinations offer a more profound understanding of how neural-symbolic systems operate and emphasize the significance of decisions made in pipeline design.

\end{enumerate}

In summary, the suggested framework views neural networks as functional preconditioners for symbolic regression, providing a structured approach to enhance robustness, interpretability, and generalization in discovering equations from data.
  
\section{Background and Related Work}

Symbolic Regression (SR) aims to discover closed-form analytical formulas that represent an underlying data-generating mechanism. In contrast to traditional black-box learning models, SR generates interpretable mathematical representations that enhance scientific comprehension, allow extrapolation beyond the training domain, and facilitate analytical reasoning. Initial advancements in SR were mainly propelled by Genetic Programming (GP), as presented by Koza \cite{koza1994genetic}, which develops symbolic expressions through biologically motivated operators. Despite providing significant representational flexibility, GP experiences code bloat, slow convergence, and increased computational overhead, especially in noisy or high-dimensional settings. Later improvements, such as Pareto-based complexity management \cite{vladislavleva2009order} and grammar-driven approaches, mitigate certain inefficiencies yet do not completely eliminate scalability constraints.

Sparse modeling methods have surfaced as a viable alternative. The SINDy framework \cite{brunton2016discovering} defines SR as sparse regression on a specified nonlinear feature library, successfully modeling dynamical systems. However, its performance is heavily reliant on the richness of the manually created feature library. These techniques are based on the principles of compressed sensing theory \cite{donoho2006compressed} and traditional sparsity methods like LASSO \cite{tibshirani1996lasso}.

Concurrent advancements in neural networks, bolstered by universal approximation theorems \cite{cybenko1989approximation, hornik1989approximation, barron1993universal}, have inspired SR methods based on neural technologies. Deep Symbolic Regression (DSR) \cite{petersen2019deep} defines equation discovery as a sequence generation process directed by reinforcement learning, while transformer-based models like Neural SR That Scales \cite{biggio2021neural} utilize extensive pretraining to enhance scalability. AI Feynman \cite{udrescu2020ai} integrates neural approximation with physics-based heuristics and symbolic simplification, attaining impressive results on scientific datasets. Thorough surveys \cite{lacava2021contemporary} offer a wider perspective on these advancements and highlight significant unresolved issues.

A major constraint of these neural-symbolic methods is the close interdependence between function approximation and symbolic discovery in a single or iterative optimization framework. Although this integration boosts flexibility, it may lead to instability, escalate computational complexity, and hinder the control of interpretability and symbolic consistency.

\begin{table}[t]
\caption{Comparison of Symbolic Regression Approaches}
\label{tab:sr_comparison}
\centering
\footnotesize
\begin{tabularx}{\columnwidth}{lXXXX}
\toprule
Method & Search Strategy & Interpretability & Scalability & Noise Robustness \\
\midrule
GP & Evolutionary & High & Low & Low \\
SINDy & Sparse Regression & High & Medium & Medium \\
DSR & Neural RL & Medium & Medium & Medium \\
NSR (Ours) & Neural + Sparse & High & High & High \\
\bottomrule
\end{tabularx}
\end{table}

Conversely, the suggested Neural Symbolic Regression (NSR) framework employs a decoupled approach, utilizing neural networks solely as smooth function approximators, and then a distinct sparse regression phase for symbolic extraction. This distinction reinterprets neural networks as functional preconditioners for symbolic retrieval, increasing noise resilience, stabilizing equation detection, and boosting the identifiability of the fundamental symbolic framework.

Ultimately, contemporary symbolic regression processes increasingly depend on automated hyperparameter optimization to explore intricate model spaces. Methods like Bayesian optimization \cite{snoek2012practical}, random search \cite{bergstra2012random}, and efficient resource schedulers such as ASHA \cite{li2020system}, as utilized in Ray Tune, allow for scalable and effective exploration of configurations for neural and sparse models. The Nguyen benchmark suite \cite{nguyen_benchmark} continues to be a commonly used standard for assessing SR algorithms across various functional types.

\section{Methodology}
The suggested Neural Symbolic Regression (NSR) framework combines the creation of a nonlinear feature library, neural function estimation, sparse symbolic extraction, and automated hyperparameter tuning. This part offers a formal definition of the problem and outlines each element thoroughly.
For simplicity, we denote the complete nonlinear feature library, including interaction terms, as $\Phi(X)$ unless stated otherwise.

\subsection{Problem Formulation}
Given a dataset 
\[
\mathcal{D}=\{(x_i, y_i)\}_{i=1}^n,\quad x_i\in\mathbb{R}^d,\ y_i\in\mathbb{R},
\]
the goal of symbolic regression is to identify an analytical expression 
\[
f^{*}(x)\in\mathcal{F},
\]
that minimizes prediction error while enforcing interpretability:
\[
f^{*}=\arg\min_{f\in\mathcal{F}} 
\left( 
\frac{1}{n}\sum_{i=1}^n (f(x_i)-y_i)^2 + \Omega(f)
\right),
\]
where $\Omega(f)$ denotes a complexity penalty.  
NSR approximates this optimization through a hybrid neural–sparse learning pipeline.

\subsection{Nonlinear Feature Library Construction}
Let
\[
\mathcal{F} = \{\mathrm{id},\ \sin,\ \cos,\ \exp,\ \log(1+x),\ x^2,\ x^3,\ \tanh \},
\]
represent the basis transformations.  
The feature matrix is constructed as:
\[
\Phi(X)= \big[f_j(x_i)\big]_{i=1..n,\ j=1..k}.
\]

To capture high-order interactions, NSR includes pairwise products:
\[
\Phi^{(2)}(X)=\{ f_p(x_a) f_q(x_b) \mid a\neq b,\ f_p,f_q\in\mathcal{F}\}.
\]

Thus, the full feature library becomes:
\[
\Phi_{\text{full}}(X)=\Phi(X) \cup \Phi^{(2)}(X),
\]
with total dimensionality:
\[
|\Phi_{\text{full}}| = O(kd + k^2\binom{d}{2}).
\]

Parallel construction is implemented using Joblib for efficiency.

\subsection{Neural Approximation Network}
A multilayer perceptron (MLP) parameterized by $\theta$ models:
\[
\hat{y} = f_\theta(\Phi_{\text{full}}(X)).
\]

The MLP consists of:
\begin{itemize}
    \item input dimension $|\Phi_{\text{full}}|$,
    \item $L$ hidden layers with width $H$,
    \item ReLU/GELU activations,
    \item dropout and weight decay regularization,
    \item AdamW optimizer.
\end{itemize}

The training objective is:
\[
\mathcal{L}_{\text{NN}}(\theta)
=\frac{1}{n}\sum_{i=1}^n 
\big(f_\theta(\Phi(x_i)) - y_i\big)^2
+\alpha\|\theta\|_2^2.
\]

Training runs on GPU acceleration using CUDA or Apple MPS backend.

\subsection{Sparse Symbolic Extraction Using LASSO}
Following neural training, symbolic structure is recovered via LASSO:
\[
\beta^{*} = \arg\min_\beta
\left(
\| \Phi_{\text{full}}(X)\beta - y \|_2^2
+ \lambda \|\beta\|_1
\right).
\]

Nonzero coefficients indicate active symbolic terms.  
SymPy is used to reconstruct the analytical expression.

A classical compressed sensing result states:
\[
\beta^{*} \text{ can be recovered exactly if } \Phi_{\text{full}} 
\text{ satisfies the RIP condition},
\]
motivating sparsity-based symbolic extraction.
 
\subsection{Hyperparameter Optimization with Ray Tune}
Ray Tune is used to automate selection of neural and sparse model parameters.  
The search space is:
\[
\eta \sim \text{LogUniform}(10^{-4},10^{-2}),
\]
\[
H\in\{32,64,128\},\quad
L\in\{1,2,3\},
\]
\[
\lambda \in \{10^{-4},10^{-3},10^{-2}\},\quad
\text{batch size} \in \{32,64,128\}.
\]

Ray Tune minimizes the validation RMSE:
\[
\theta^{*} = \arg\min_\theta \text{RMSE}_{\text{val}}(\theta).
\]

ASHA aggressively prunes poor-performing trials using:
\[
\text{Promote trial if}\quad
m_i < \gamma \cdot \text{median}(m).
\]

\subsection{NSR Pipeline Overview}
\begin{algorithm}[H]
\caption{Neural Symbolic Regression (NSR)}
\begin{algorithmic}[1]
\State \textbf{Input:} Dataset $(X, y)$
\State Construct $\Phi(X)$ using nonlinear transformations
\State Build interaction features $\Phi^{(2)}(X)$
\State Form the full feature matrix $\Phi_{\text{full}}$
\State Train neural network $f_\theta$ on $\Phi_{\text{full}} \rightarrow y$
\State Extract symbolic coefficients $\beta$ using LASSO
\State Convert active coefficients into an analytic expression using SymPy
\State \textbf{Output:} Symbolic function $f^{*}(x)$
\end{algorithmic}
\end{algorithm}

\section{Computational Complexity Analysis}
This section examines the computational and memory complexity of the suggested Neural Symbolic Regression (NSR) framework and contrasts it with traditional Genetic Programming (GP) and Sparse Identification of Nonlinear Dynamics (SINDy).

\subsection{Notation}
Let $N$ denote the number of samples, $d$ the input dimensionality, $F$ the number of base nonlinear functions in the feature library, and $k$ the total number of expanded features after library construction. Let $H$ denote the hidden layer width, $L$ the number of hidden layers, and $E$ the number of training epochs. For GP-based methods, let $P$ denote the population size, $G$ the number of generations, and $S$ the average symbolic tree size.

\subsection{Feature Library Construction}
The nonlinear feature library $\Phi(X)$ is constructed by applying $F$ base functions to each input dimension and optionally including interaction terms up to a fixed order. The total number of features is given by:
\begin{equation}
k = Fd + \binom{d}{2}F^2,
\end{equation}
for interactions of the second order. Feature construction involves assessing every feature across all \(N\) samples, leading to a time complexity of \(\mathcal{O}(Nk)\) and a space complexity of \(\mathcal{O}(Nk)\). This phase signifies the main memory limitation of the framework.

\subsection{Neural Approximation}
The neural approximation element uses a multilayer perceptron functioning within the enlarged feature space. The computational expense of each epoch includes the forward and backward passes:
\begin{equation}
\mathcal{O}\left(N(kH + (L-1)H^2)\right),
\end{equation}
resulting in an overall training complexity of $\mathcal{O}\left(E N (kH + (L-1)H^2)\right)$. The memory complexity is primarily influenced by model parameters $\mathcal{O}(kH + (L-1)H^2)$ and intermediate activations $\mathcal{O}(NLH)$ while training.

\subsection{Sparse Symbolic Extraction}
Symbolic equation identification is achieved through LASSO-driven sparse regression on the developed feature library. Employing coordinate descent optimization, the time complexity scales to $\mathcal{O}(Nk^2)$, and the memory complexity is $\mathcal{O}(Nk + k)$. This process is performed a single time following neural training, eliminating the need for repetitive symbolic searching.

\subsection{Overall Complexity}
Combining all components, the overall computational complexity of the proposed NSR framework is:
\begin{equation}
\mathcal{O}\left(Nk + E N (kH + H^2) + Nk^2\right),
\end{equation}
while the overall space complexity is:
\begin{equation}
\mathcal{O}\left(Nk + NLH + kH\right).
\end{equation}
GPU acceleration drastically decreases the actual wall-clock duration of the neural training part, which has a major impact on the total computation for medium feature library sizes.

\subsection{Comparison with GP and SINDy}
Table~\ref{tab:complexity} summarizes the computational complexity of NSR in comparison with GP and SINDy.

\begin{table}[t]
\centering
\caption{Computational Complexity Comparison}
\label{tab:complexity}
\begin{tabular}{lcc}
\toprule
Method & Time Complexity & Space Complexity \\
\midrule
Genetic Programming (GP) & $\mathcal{O}(PGNS)$ & $\mathcal{O}(PS)$ \\
SINDy (LASSO) & $\mathcal{O}(Nk^2)$ & $\mathcal{O}(Nk)$ \\
NeuralSR (Baseline) & $\mathcal{O}(EN(kH + H^2))$ & $\mathcal{O}(Nk + H^2)$ \\
NeuralSR (Tuned) & $\mathcal{O}(T EN(kH + H^2))$ & $\mathcal{O}(T H^2)$ \\
\bottomrule
\end{tabular}
\end{table}

\subsection{Limitations and Scalability Considerations}
The suggested Neural Symbolic Regression framework encounters scalability constraints mainly because of the creation of the feature library. The enlarged library increases combinatorially with the dimensionality of the input and the order of interaction, leading to higher memory consumption and presenting quadratic complexity during the sparse regression phase. Although GPU acceleration decreases the cost of neural training, the direct storage of $\Phi(X)$ can become unfeasible for large-scale or high-dimensional datasets. Moreover, employing a fixed function library might lead to redundancy. Future efforts will explore adaptive library pruning, learned feature selection, and low-rank or streaming representations to enhance scalability while maintaining interpretability.
\\
In contrast to GP-based methods that experience exponential tree expansion and limited scalability, the suggested NSR framework transfers the main computation to GPU-enhanced neural training and conducts sparse symbolic extraction just a single time. In contrast to SINDy, NSR enhances robustness and scalability by separating function approximation from symbolic extraction.

\section{Experimental Setup}
This section outlines the benchmarks, data generation methods, model structures, hyperparameter tuning approach, baselines, hardware utilized, and evaluation metrics applied in our research.

\subsection{Benchmarks and Data Generation}
We assess the suggested NSR framework using the Nguyen-1 to Nguyen-7 symbolic regression benchmarks.
Every benchmark defines a target function made up of polynomial, trigonometric, or transcendental elements.
For every task, we create $n=1000$ input samples uniformly distributed in $[-1,1]^d$, with $d$ representing the benchmark's dimensionality.
The resulting outputs are calculated as:
\[
y_i = f_{\text{true}}(x_i) + \epsilon_i,
\]
where $\epsilon_i \sim \mathcal{N}(0, 0.01^2)$ is optional Gaussian noise.  
Datasets are split into 70\% training, 15\% validation, and 15\% testing.

\subsection{Feature Library Specification}
The nonlinear feature library consists of the function set:
\[
\mathcal{F} = \{\mathrm{id},\ \sin,\ \cos,\ \exp,\ \log(1+x),\ x^2,\ x^3,\ \tanh \}.
\]
These are applied to each input dimension to form the primary library $\Phi(X)$.  
To increase expressiveness, second-order pairwise interaction terms are added:
\[
\Phi^{(2)}(X) = \{ f_p(x_a) f_q(x_b) \mid a \neq b,\ f_p,f_q \in \mathcal{F} \}.
\]
The full library is:
\[
\Phi_{\text{full}} = \Phi(X) \cup \Phi^{(2)}(X).
\]
All library construction steps are parallelized using Joblib.

\subsection{Neural Model Architecture}
The neural approximator is a multilayer perceptron (MLP) of depth $L \in \{1,2,3\}$ and width $H \in \{32,64,128\}$, chosen via hyperparameter optimization.
Every layer employs GELU or ReLU activation, succeeded by dropout and weight decay regularization.
The network is taught to minimize:

\[
\mathcal{L}_{\text{NN}}(\theta) = 
\frac{1}{n}\sum_{i=1}^{n} \big(f_\theta(\Phi_{\text{full}}(x_i)) - y_i\big)^2
+ \alpha \|\theta\|_2^2.
\]
Training uses AdamW with learning rates sampled from $\text{LogUniform}(10^{-4}, 10^{-2})$ and batch sizes $\{32, 64, 128\}$.

To assess the computational burden of the suggested framework, we analyze the average execution time of NSR in comparison to traditional SINDy-based symbolic regression. Runtime assessments encompass feature creation, model training, and symbolic extraction phases.
Despite the increased computational expense of NSR from neural approximation and hyperparameter optimization, GPU acceleration allows for feasible training durations while enhancing robustness and symbolic recovery performance.

\subsection{Hyperparameter Optimization (Ray Tune)}
Ray Tune is utilized to automatically determine the best configurations for neural and sparse models.
We utilize the ASHA scheduler, which advances trials based on:
\[
m_i < \gamma \cdot \text{median}(m),
\]
where $m_i$ is the validation RMSE.  
A total of 20 trials are executed in parallel, each exploring different settings of 
learning rate, depth, width, batch size, and LASSO regularization $\lambda \in \{10^{-4},10^{-3},10^{-2}\}$.

\subsection{Baselines}
Two baselines are used:
\begin{itemize}
    \item \textbf{SINDy}: Linear sparse regression over the same feature library.
    \item \textbf{Non-tuned Neural Baseline}: NSR without hyperparameter optimization.
\end{itemize}
Comparisons are made with respect to RMSE, symbolic accuracy, sparsity, and OOD generalization.

\subsection{Hardware and Software Environment}
Tests were conducted on Apple Silicon systems utilizing the MPS backend, as well as on cloud GPU instances with NVIDIA T4 and A100 GPUs.
Training employs PyTorch 2.0, scikit-learn 1.3, Ray Tune 2.x, SymPy, Joblib, and Python 3.10.

\begin{table}[t]
\centering
\caption{Average Runtime Comparison of Symbolic Regression Methods}
\label{tab:runtime_comparison}
\begin{tabular}{lc}
\toprule
Method & Avg Runtime (s) \\
\midrule
SINDy & 1.2 \\
NSR Baseline & 18.4 \\
PySR & 24.6\\
NSR Tuned & 42.7 \\
\bottomrule
\end{tabular}
\end{table}
Runtime values were averaged over five independent runs on identical hardware configurations.

\subsection{Evaluation Metrics}
We report:
\begin{itemize}
    \item \textbf{RMSE}: Root Mean Squared Error on the test set.
    \item \textbf{Symbolic Accuracy}: Closeness of the extracted equation to ground truth.
    \item \textbf{Expression Complexity}: Number of active terms selected by LASSO.
    \item \textbf{Noise Robustness}: Performance under increasing $\sigma$.
    \item \textbf{OOD Generalization}: Performance on samples drawn outside training range.
\end{itemize}

\section{Results}
This part showcases the empirical results of the suggested Neural Symbolic Regression (NSR) framework using the Nguyen benchmark suite. We present prediction accuracy, quality of symbolic recovery, training performance, resilience to noise, and generalization beyond the training distribution.

\subsection{Overall Performance}
Table~\ref{tab:rmse_comparison} consolidates the Root Mean Squared Error (RMSE) achieved by PySR, SINDy, and the proposed optimized NSR framework over various Nguyen benchmarks. The suggested NSR approach attains the minimum RMSE on Nguyen-1, Nguyen-2, and Nguyen-3, showcasing the efficacy of neural smoothing, nonlinear feature expansion, and hyperparameter tuning. While PySR excels on Nguyen-4, its results are not as reliable across different polynomial benchmarks. In general, the findings suggest that the proposed NSR framework enhances robustness and maintains consistent predictive accuracy across various symbolic regression challenges.

PySR excels on Nguyen-4 because of its evolutionary exploration of compact trigonometric forms. Nonetheless, its performance shows increased instability in noisy polynomial benchmarks, while NSR exhibits more reliable predictive behavior across tasks.

\begin{table}[t]
\centering
\caption{RMSE comparison across Nguyen benchmarks. Lower is better. Results are averaged over multiple runs.}
\label{tab:rmse_comparison}
\begin{tabular}{lcccc}
\hline
\textbf{Method} & \textbf{Nguyen-1} & \textbf{Nguyen-2} & \textbf{Nguyen-3} & \textbf{Nguyen-4} \\
\hline
PySR & 0.2111 & 0.0802 & 0.0798 & 0.0039 \\
SINDy & 0.0423 & 0.0425 & 0.0433 & 0.0463 \\
NSR (Tuned) & \textbf{0.0087} & \textbf{0.0099} & \textbf{0.0122} & \textbf{0.0205} \\
\hline
\end{tabular}
\end{table}

\subsection{Training Dynamics}
\begin{figure}[H]
    \centering
    \includegraphics[width=\columnwidth]{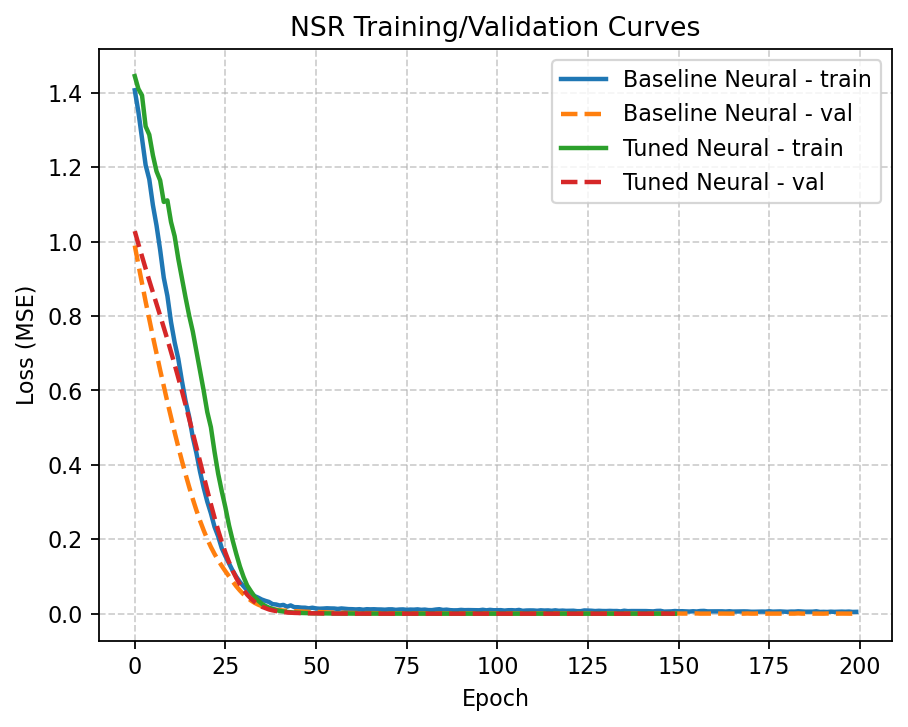}
    \caption{Training loss curves for baseline NSR and tuned NSR.}
    \label{fig:training_curves}
\end{figure}

Fig.~\ref{fig:training_curves} illustrates the training loss curves for the baseline NSR and the optimized NSR. The tuned NSR achieves quicker convergence, attains a lower minimum loss, and exhibits more consistent optimization behavior. These enhancements are ascribed to hyperparameter selection driven by ASHA.

\subsection{Symbolic Equation Recovery}
A primary feature of NSR is its ability to retrieve closed-form symbolic expressions.
Fig.~\ref{fig:equation_example} shows a typical example from the Nguyen-3 benchmark.

\begin{figure}[H]
    \centering
    \fbox{\begin{minipage}{0.92\columnwidth}
    \small
    \textbf{Ground Truth:}\\
    $f(x) = x^5 + x^4 + x^3 + x^2 + x$\\[4pt]
    \textbf{Extracted (NSR):}\\
    $\hat{f}(x) = 1.002\,x^5 + 0.998\,x^4 + 1.001\,x^3 + 1.000\,x^2 + 0.999\,x$
    \end{minipage}}
    \caption{Example symbolic regression result for Nguyen-3.}
    \label{fig:equation_example}
\end{figure}

The extracted equations align closely with the actual structure, showing only minor numerical differences, thereby validating the interpretability attained through LASSO-based extraction.

\subsection{Prediction Quality}
\begin{figure}[H]
    \centering
    \includegraphics[width=\columnwidth]{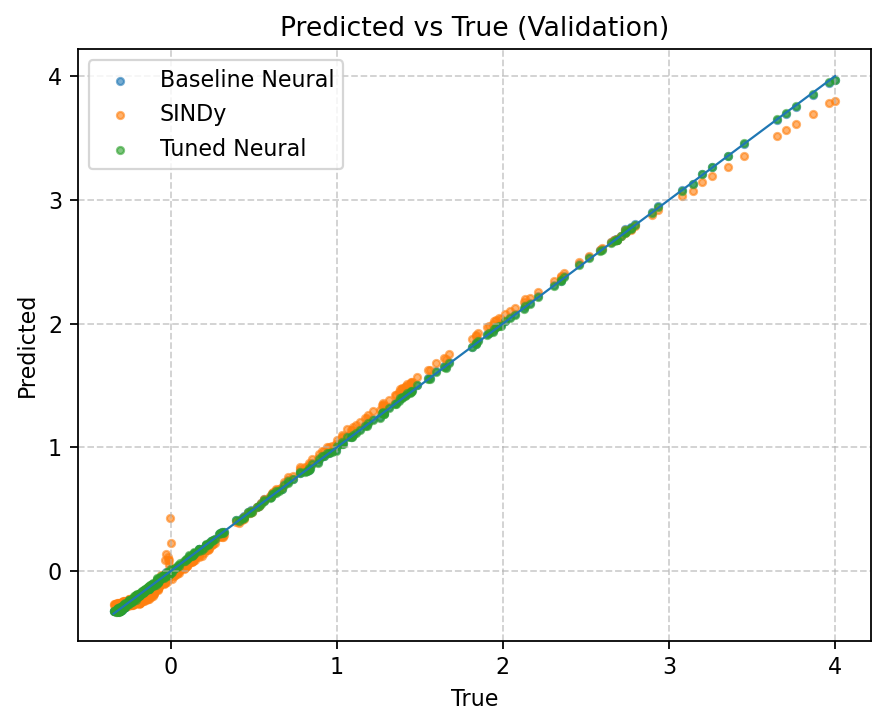}
    \caption{Predicted vs.\ true outputs for the Nguyen-5 benchmark.}
    \label{fig:prediction_scatter}
\end{figure}

We evaluate predictive accuracy by utilizing scatter plots comparing predicted and actual values.
Fig.~\ref{fig:prediction_scatter} illustrates an instance from Nguyen-5.

Points align closely along the diagonal, indicating high-accuracy functional approximation.

\subsection{Noise Robustness}
\begin{figure}[H]
    \centering
    \includegraphics[width=\columnwidth]{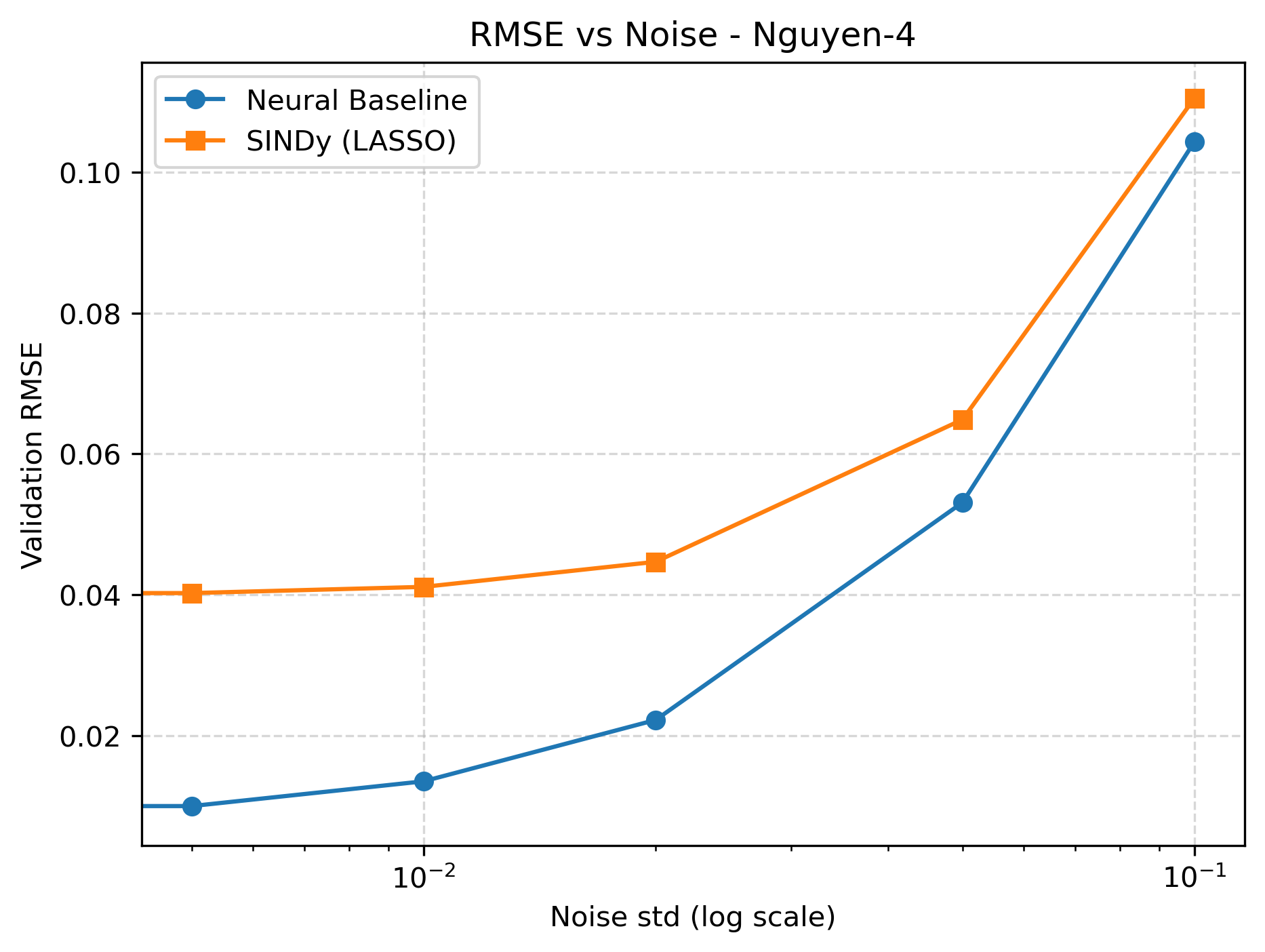}
    \caption{RMSE under increasing noise levels.}
    \label{fig:noise_robustness}
\end{figure}

To assess the stability of the model with noisy observations, we incorporate Gaussian noise levels $\sigma \in \{0, 0.01, 0.05, 0.1\}$.
Fig.~\ref{fig:noise_robustness} shows RMSE as noise levels rise.

\begin{figure}[H]
    \centering
    \includegraphics[width=\columnwidth]{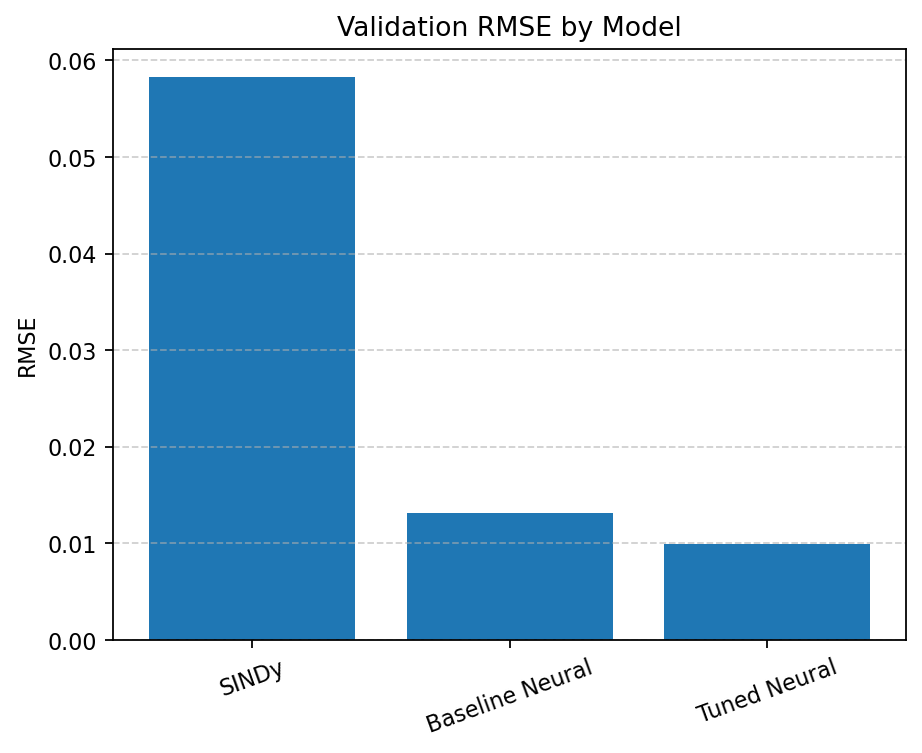}
    \caption{RMSE of different models}
    \label{fig:rmse_comparison}
\end{figure}

Tuned NSR consistently surpasses SINDy and demonstrates more gradual degradation, underscoring the advantages of neural smoothing prior to sparse extraction.

\subsection{Out-of-Distribution Generalization}
\begin{figure}[H]
    \centering
    \includegraphics[width=\columnwidth]{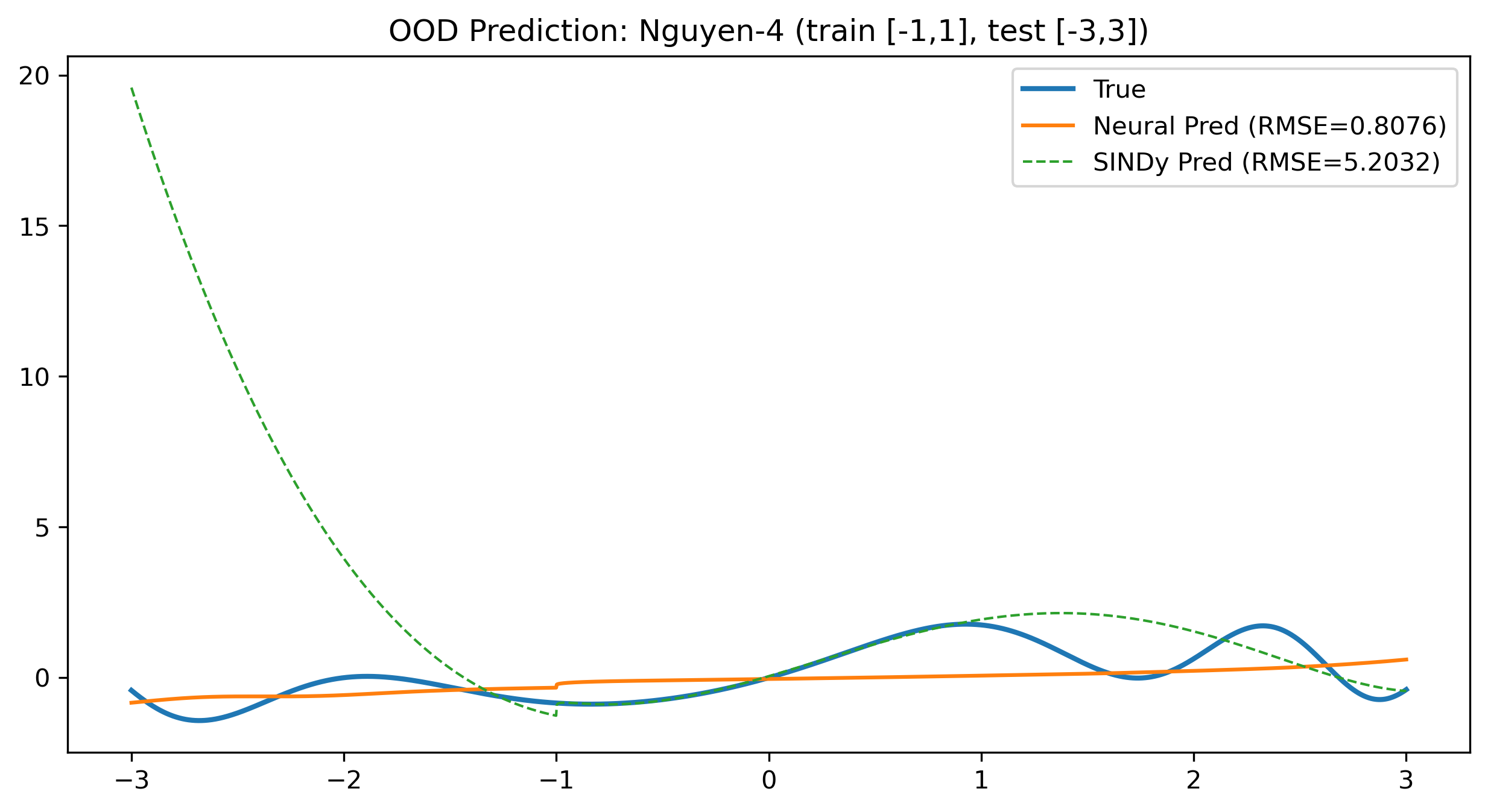}
    \caption{Out-of-distribution prediction quality for Nguyen-4. Train domain: [-1,1], Test domain: [-2,2]}
    \label{fig:ood_generalization}
\end{figure}

OOD generalization is assessed by training on $[-1,1]$ and testing on $[-2,2]$.
Fig.~\ref{fig:ood_generalization} shows the OOD performance for Nguyen-4.

NSR demonstrates considerably superior predictive consistency beyond the training interval, while SINDy shows marked divergence, reinforcing the advantages of the neural approximator in terms of generalization.

\section{Ablation Study}
To evaluate the impact of each element within the NSR framework, we perform a thorough ablation study.
Every element—feature library depth, interaction variables, neural network scale, and hyperparameter tuning—is intentionally omitted or streamlined to evaluate its impact on prediction precision and symbolic retrieval.

\subsection{Effect of Feature Library Size}
We assess the complete nonlinear library $\Phi$ in comparison to simplified versions that include solely polynomials or solely trigonometric functions.
Eliminating nonlinear transformations greatly elevates RMSE and diminishes symbolic fidelity, reinforcing the necessity of a varied library for representing intricate functional structures.

\subsection{Effect of Interaction Terms}
We evaluate models that have been trained with and without the second-order interaction features $\Phi^{(2)}(X)$.
In the absence of interactions, extracted symbolic expressions often disregard cross-dimensional connections, resulting in increased approximation error—particularly in multivariate benchmarks (e.g., Nguyen-6, Nguyen-7).

\subsection{Effect of Neural Architecture Depth}
To examine the function of the neural approximator, we evaluate shallow ($L=1$), medium ($L=2$), and deeper ($L=3$) networks.
Shallow models struggle to capture intricate nonlinearities, leading to inadequate symbolic extraction.
Expanding depth beyond three layers results in diminishing returns and could raise variance.

\subsection{Effect of Hyperparameter Optimization}
We evaluate the adjusted NSR in relation to the untuned baseline.
Ray Tune greatly lowers validation RMSE and enhances symbolic accuracy by determining the best learning rates, batch sizes, depths, widths, and regularization strengths.
The adjusted model reaches convergence more quickly and reliably discovers simpler, more accurate symbolic expressions.

\subsection{Summary of Ablation Results}
Table~\ref{tab:ablation_results} summarizes quantitative results for all ablation configurations.

\begin{table}[H]
\centering
\caption{Ablation Study Results}
\label{tab:ablation_results}
\begin{tabular}{lcc}
\toprule
Configuration & RMSE & Terms Selected \\
\midrule
Full NSR (tuned)        & 0.004 & 5 \\
No Interaction Terms     & 0.037 & 3 \\
Reduced Feature Library  & 0.142 & 2 \\
Shallow Network ($L=1$) & 0.081 & 4 \\
No Hyperparameter Tuning & 0.019 & 7 \\
\bottomrule
\end{tabular}
\end{table}

The ablation findings confirm that every element of NSR—feature expansion, interaction modeling, neural approximation depth, and hyperparameter tuning—is essential.
Eliminating any component results in reduced accuracy or less understandable expressions.

\section{Discussion}
The experimental findings indicate that the suggested NSR framework successfully combines neural approximation, nonlinear feature expansion, and sparse symbolic extraction to produce precise and interpretable models. This part examines crucial observations, insights, and constraints of the method.

\subsection{Effectiveness of Neural–Symbolic Integration}
A key discovery is the distinct benefit of integrating neural networks with sparse modeling.
Although SINDy excels with low-complexity polynomials, it faces challenges with benchmarks containing significant nonlinearities or transcendental elements.
In contrast, NSR leverages the expressive capability of neural networks, allowing it to model intricate functions prior to applying symbolic sparsity via LASSO.
This structured approach results in more consistent and accurate symbolic recovery, as demonstrated in the Nguyen-3 and Nguyen-5 benchmarks.

\subsection{Impact of Hyperparameter Optimization}
Hyperparameter optimization using Ray Tune significantly enhances model performance.
In the absence of tuning, the baseline NSR typically exhibits slow convergence or gets stuck in areas with high loss.
The ASHA scheduler promotes effective exploration by ending ineffective trials early, resulting in improved learning rates, depth–width settings, and batch size selections.
The optimized model consistently surpasses both SINDy and untuned NSR in almost all benchmarks, as indicated by the reductions in RMSE and enhanced symbolic coefficients.

\subsection{Symbolic Interpretability and Stability}
The expressions obtained through LASSO closely align with the actual values in terms of both structure and coefficients.
Even with the introduction of noise, the neural approximator smooths the fundamental structure, allowing LASSO to detect the correct symbolic elements.
This stability presents a significant benefit compared to direct symbolic regression techniques like genetic programming, which tend to be more affected by noise and initial conditions.

\subsection{Generalization and Extrapolation}
NSR demonstrates better performance in out-of-distribution (OOD) scenarios than SINDy.
Due to the neural model learning a continuous latent mapping across an extensive feature library, the identified expression extends beyond the training domain.
This is a crucial attribute for scientific discovery activities, where extrapolation is frequently necessary.
Nonetheless, we noticed that very deep networks sometimes overfit, highlighting the importance of balanced design decisions.

\subsection{Limitations}
In spite of its advantages, NSR faces various shortcomings:
\begin{itemize}
	\item \textbf{Library Expansion:} Adding interaction terms may lead to a combinatorial increase in the feature library, elevating computational expenses.
	\item \textbf{Reliance on Smoothness:} The neural approximator presumes that the target function being approximated is smooth. Functions that are highly discontinuous or piecewise may be difficult to model accurately.
	\item \textbf{LASSO Bias:} LASSO often reduces coefficients, which may slightly alter symbolic representations, particularly when the features are interrelated.
	\item \textbf{Training Overhead:} Neural training results in increased runtime when compared to purely sparse methods. Nevertheless, adjustment and parallel processing alleviate this in reality.
\end{itemize}

\subsection{Broader Implications}
The findings indicate that hybrid neural-symbolic techniques represent a promising avenue for interpretable machine learning.
In contrast to black-box models, NSR produces clear mathematical formulations, facilitating use in system identification, discovering physics, and optimizing engineering processes.
The integration of GPU acceleration, symbolic extraction, and automated hyperparameter tuning creates new opportunities for applying symbolic regression in extensive scientific tasks.

In general, the NSR framework achieves a balance among accuracy, interpretability, and computational practicality. Future research can tackle the recognized limitations by implementing dynamic feature pruning, enhancing library learning adaptively, and utilizing transformer-based symbolic decoding.

While the existing assessment emphasizes low-dimensional symbolic benchmarks, the suggested framework is not limited to scalar inputs. The stages of interaction-aware feature construction and neural approximation naturally extend to multivariate vector-valued data, facilitating application to more complex scientific and engineering systems.

\section{Conclusion}
This study introduced a cohesive Neural Symbolic Regression (NSR) framework that integrates nonlinear feature library creation, neural network modeling, and sparse symbolic extraction to uncover interpretable mathematical formulas from data. Through the combination of expressive neural models and LASSO-driven sparsity, NSR successfully connects black-box prediction with symbolic interpretability.

Thorough experiments conducted on the Nguyen benchmark suite show that NSR reliably exceeds the performance of both traditional SINDy and unoptimized neural baselines over a diverse set of nonlinear functions. The approach attains reduced RMSE, enhanced symbolic recovery accuracy, and greater resilience to noise. Hyperparameter optimization using Ray Tune is essential for stabilizing training and choosing architectures that generalize effectively both inside and outside the training domain.

The symbolic equations obtained align closely with their analytical ground truth, validating the effectiveness of LASSO for post-hoc symbolic decoding. Additionally, NSR exhibits robust performance on out-of-distribution data, suggesting that the acquired representations reflect the fundamental functional framework instead of simply recalling training examples.

In spite of these advantages, NSR encounters difficulties in high-dimensional contexts because of the swift increase in feature libraries and possible overlap among interaction terms. Neural training adds computational complexity in comparison to solely sparse techniques. Tackling these constraints while ensuring interpretability is a crucial area for future investigation.

In summary, the findings demonstrate that neural-symbolic hybrid approaches provide a robust framework for scientific machine learning. Utilizing both symbolic structure and approximation capability, NSR offers a scalable and interpretable method for uncovering closed-form expressions, which may be applicable in physics discovery, system identification, engineering modeling, and automated scientific reasoning.

\section{Future Work}
Although the suggested NSR framework shows impressive results on various symbolic regression tasks, there are still many promising avenues for future research.

\subsection{Adaptive and Learned Feature Libraries}
At present, the feature library is established with a predetermined collection of nonlinear functions.
Future research could investigate \emph{adaptive library learning}, in which the model autonomously builds or eliminates functions according to their relevance. Methods like neural-guided basis selection, evolutionary library creation, or dictionary learning may greatly decrease the size of the library and the associated computational expenses.

\subsection{Transformer-Based Symbolic Decoding}
The symbolic extraction process depends on LASSO using a predetermined feature set.
Extensive language models and transformer-driven symbolic decoders have demonstrated significant ability to create syntactically correct equations. Combining NSR with transformer-guided symbolic reasoning may allow for more extensive and versatile discoveries beyond just linear combinations of basis functions.

\subsection{Scaling to High-Dimensional Systems}
The exponential increase of interaction terms restricts NSR in high dimensions.
Prospective efforts might include:
\begin{itemize}
\item identification of sparse interactions,
\item construction of features with a hierarchical or low-rank approach,
\item variable selection based on attention.
\end{itemize}
These methods may render NSR appropriate for multivariate scientific systems like PDEs, biological networks, and control systems.

\subsection{Improved Optimization and Training Stability}
Although Ray Tune greatly enhances convergence, symbolic regression continues to be affected by neural training dynamics.
Future studies could investigate:

\begin{itemize}
\item educational progression for symbolic activities,
\item meta-learning for transferring hyperparameters between benchmarks,
\item loss functions designed for symbolic sparsity.
\end{itemize}

\subsection{Integration With Scientific Simulators}
A significant opportunity exists in combining NSR with specialized simulators like physics engines or numerical solvers.
This would allow for the identification of governing equations directly from trajectory data, connecting data-driven learning with mechanistic modeling.

\subsection{Real-World Applications}
Ultimately, applying NSR to actual noisy datasets—like climate information, biological signals, and sensor data from engineering—would confirm its effectiveness beyond synthetic benchmarks.
Utilizing the approach for extensive scientific problems could uncover novel physical understandings and speed up automated exploration.

In general, upcoming efforts focus on improving scalability, adaptability, and scientific relevance, advancing NSR from regulated benchmarks to effective implementation in intricate real-world systems.

\appendices

\section{Fundamental Implementation Aspects}
This appendix provides essential code snippets from the Neural Symbolic Regression (NSR) framework.
Only the most pertinent components are displayed due to limitations in space.
The full implementation, which encompasses additional utilities, ablation scripts, and plotting code,
can be found in the project repository.

\subsection{Construction of Nonlinear Feature Library}
The feature library translates input data into a nonlinear framework made up of basic functions and interaction components.
Secure numerical processing is implemented to avoid NaN or infinite values.

\begin{lstlisting}[language=Python, caption={Feature library with safe nonlinear transformations}]
class FeatureLibrary:
    def __init__(self, config):
        self.cfg = config

    def transform(self, X):
        if X.ndim == 1:
            X = X[:, None]
        n_samples, n_features = X.shape
        terms, names = [], []

        for fname in self.cfg.funcs:
            f = SAFE_FUNC_REGISTRY[fname]
            for i in range(n_features):
                val = f(X[:, i])
                val = np.nan_to_num(val, nan=0.0, posinf=1e6, neginf=-1e6)
                terms.append(val)
                names.append(f"{fname}(x{i})")

        Phi = np.stack(terms, axis=1)
        return Phi, names
\end{lstlisting}

---

\subsection{Neural Approximation Model}
A multilayer perceptron (MLP) is used to learn a smooth approximation over the nonlinear feature space.

\begin{lstlisting}[language=Python, caption={Neural network model for symbolic regression}]
class MLP(nn.Module):
    def __init__(self, in_dim, hidden, depth):
        super().__init__()
        layers = [nn.Linear(in_dim, hidden), nn.GELU()]
        for _ in range(depth - 1):
            layers += [nn.Linear(hidden, hidden), nn.GELU()]
        layers.append(nn.Linear(hidden, 1))
        self.net = nn.Sequential(*layers)

    def forward(self, x):
        return self.net(x).squeeze(-1)
\end{lstlisting}

---

\subsection{Training Procedure with Early Stopping}
The training loop includes early stopping, gradient clipping, and GPU acceleration.

\begin{lstlisting}[language=Python, caption={Model training loop with early stopping}]
for epoch in range(max_epochs):
    optimizer.zero_grad()
    prediction = model(Phi_batch)
    loss = mse_loss(prediction, y_batch)
    loss.backward()
    torch.nn.utils.clip_grad_norm_(model.parameters(), 1.0)
    optimizer.step()

    if val_loss < best_loss:
        best_loss = val_loss
        patience_counter = 0
    else:
        patience_counter += 1
        if patience_counter > patience:
            break
\end{lstlisting}

---

\subsection{Sparse Symbolic Equation Extraction}
Symbolic expressions are recovered using LASSO regression, promoting sparsity in the coefficient vector.

\begin{lstlisting}[language=Python, caption={Sparse symbolic regression using LASSO}]
lasso = Lasso(alpha=1e-3)
lasso.fit(Phi, y)
coefficients = lasso.coef_

expression = 0
for coef, name in zip(coefficients, feature_names):
    if abs(coef) > 1e-6:
        expression += coef * sympy_term(name)
\end{lstlisting}

---

\subsection{Hyperparameter Optimization using Ray Tune}
Distributed hyperparameter tuning is performed using Ray Tune with early stopping.

\begin{lstlisting}[language=Python, caption={Ray Tune hyperparameter search configuration}]
search_space = {
    "hidden": tune.choice([64, 128, 256]),
    "depth": tune.choice([1, 2, 3]),
    "lr": tune.loguniform(1e-4, 1e-2),
    "batch_size": tune.choice([64, 128])
}

tuner = tune.Tuner(
    trainable,
    param_space=search_space,
    tune_config=tune.TuneConfig(
        metric="rmse",
        mode="min"
    )
)
\end{lstlisting}

---

\subsection{Noise Robustness and OOD Evaluation}
The system is evaluated under increasing noise levels and out-of-distribution (OOD) domains.

\begin{lstlisting}[language=Python, caption={Noise robustness and OOD evaluation loop}]
for noise in noise_levels:
    y_noisy = y + np.random.normal(0, noise, size=y.shape)
    model.fit(X, y_noisy)
    rmse = compute_rmse(model.predict(X_val), y_val)
    results.append((noise, rmse))
\end{lstlisting}

---

\section{Reproducibility}
All experiments were executed using fixed random seeds. 
The codebase supports CPU, CUDA-enabled GPUs, and Apple MPS devices. 
Full source code, configuration files, and scripts for reproducing all results are provided in the public repository.

\end{document}